\documentclass[smallcondensed,envcountsect]{svjour3} 
\usepackage{amsmath, amssymb}
\usepackage{multirow}

\smartqed  % flush right qed marks, e.g. at end of proof
\usepackage{graphicx}
\usepackage{color}
\usepackage[colorlinks=true,linkcolor=red,citecolor=blue,urlcolor=cyan]{hyperref}
\usepackage{algorithm}
\usepackage[noend]{algpseudocode}
\usepackage{epsfig}
\usepackage{enumitem}

\def\makeheadbox{\relax}

\usepackage{centernot}
\usepackage{mathtools}
\usepackage{mathrsfs}
\DeclareMathOperator*{\argmax}{arg\,max}

\spnewtheorem{thm}{Theorem}[section]{\bfseries}{\itshape}
\newcommand{\ie}{{\it i.e.}}
\newcommand{\eg}{{\it e.g.}}

\begin{document}

\title{A non-autonomous equation discovery method for time signal classification\thanks{B. Osting acknowledges partial support from NSF DMS 17-52202.  H. S. Bhat acknowledges partial support from NSF DMS 17-23272.}}

\titlerunning{NAED method for time signal classification}

% \subjclass[2010]{34H05, % Control problems involving ordinary differential equations
% 68T07, % Artificial neural networks and deep learning
% 62L10. % Sequential statistical analysis
% }

% \keywords{Time signal analysis; classification; equation discovery; neural networks; adjoint method}

\author{Ryeongkyung Yoon \and Harish S. Bhat \and Braxton Osting}

\authorrunning{R. Yoon, H. S. Bhat, and B. Osting} % if too long for running head

\institute{R. Yoon \at
              Department of Mathematics, University of Utah, Salt Lake City, UT \\
              \email{rkyoon@math.utah.edu}      
           \and
           H.~S. Bhat \at
           Department of Applied Mathematics, University of California, Merced, CA \\
           \email{hbhat@ucmerced.edu} 
           \and
           B. Osting \at 
           Department of Mathematics, University of Utah, Salt Lake City, UT \\
           \email{osting@math.utah.edu}
}

%\author{Ryeongkyung Yoon}
%\address{Department of Mathematics, University of Utah, Salt Lake City, UT}
%\email{rkyoon@math.utah.edu}

%\author{Harish S. Bhat}
%\address{Department of Applied Mathematics, University of California, Merced, CA}
%\email{hbhat@ucmerced.edu}

%\author{Braxton Osting}
%\address{Department of Mathematics, University of Utah, Salt Lake City, UT}
%\email{osting@math.utah.edu}

\date{\today}

\maketitle

%------
\begin{abstract}
Certain neural network architectures, in the infinite-layer limit, lead to systems of nonlinear differential equations.  Motivated by this idea, we develop a framework for analyzing time signals based on non-autonomous dynamical equations. We view the time signal as a forcing function for a dynamical system that governs a time-evolving hidden variable. As in equation discovery, the dynamical system is represented using a dictionary of functions and the coefficients are learned from data. This framework is applied to the time signal classification problem.  We show how gradients can be efficiently computed using the adjoint method, and we apply methods from dynamical systems to establish stability of the classifier. Through a variety of experiments, on both synthetic and real datasets, we show that the proposed method uses orders of magnitude fewer parameters than competing methods, while achieving comparable accuracy.  We created the synthetic datasets using dynamical systems of increasing complexity; though the ground truth vector fields are often polynomials, we find consistently that a Fourier dictionary yields the best results.  We also demonstrate how the proposed method yields graphical interpretability in the form of phase portraits.
 % Sequential statistical analysis
\end{abstract}

% \tableofcontents

% arise in a broad spectrum of applications and, 
%  We develop a framework based on non-autonomous dynamical systems for processing time signals, \ie, continuous-time, vector-valued data. We use this framework to develop and implement a method for the time signal classification problem. 

%---------------------------------------------
\section{Introduction}\label{sec1}

Time signals, due to their temporal ordering, multiple scales, high dimension, and autocorrelation, require special tools for meaningful analysis. In this work, we propose a non-autonomous dynamical systems framework to address \emph{time signal classification}, the problem of learning a mapping that assigns a distribution over labels
$y \in \mathbb R^{|\mathcal Y|}$ 
to a vector-valued, continuous-time signal 
$x\colon [0,T] \to  \mathbb R^n$. 
Here $\mathcal Y$ is a finite set of labels. We study the proposed method both theoretically and empirically.  We find that, with its principled and parsimonious dictionary representation of the dynamical system's vector field, the proposed method approaches and/or exceeds the test set accuracy of competing methods.

% We have a principled choice of dictionary.  We have some theory.
% The dictionary need not match the ground truth terms.  No trigonometric vector field.
% We never use that many parameters.
% We do just as well as the NCDE method and more common RNN methods.

%Let $N$ be the sample size and let $[N] :=\{1,2,\ldots,N\}$. We assume that we are given data 
%$ \left\{x_i(t), \ T_i, \  y_i \right\}_{i \in [N]}$,
%where the $i$-th time signal $x_i\colon [0,T_i] \to  \mathbb R^n$ has length $T_i$ and corresponding label $y_i \in \mathcal Y$.  

% Over the past few decades, many methods---both unsupervised and supervised---have been developed to perform a variety of time signal analysis tasks, including prediction and forecasting, classification, segmentation, and denoising.

%Note that we allow for the possibility that the time signals have different lengths. 

%We refer to the case where $|\mathcal Y| = 2$ as binary classification, but generally do not make this assumption. 

We distinguish between time signals and time series; time signals are continuous in time, while time series are discrete in time. Frequently time series are obtained from sampling a time signal at discrete times. 
Applications of time signal classification include 
predicting the genre of music based on a sound recording \cite{genre}, 
recognizing human activity using mobile sensors \cite{humanactivity}, 
diagnosing disease based on electrical biosignals (\eg, EEG, ECG, and EMG) \cite{EEG,ECG,EMG}, 
detecting natural phenomena such as earthquakes or volcanic eruptions using geophysical signals \cite{geophysics}, and automatically distinguishing between mosquito species using wing-beat recordings \cite{wingbeat}.

One of the most promising approaches to time series classification involves deep recurrent neural networks (RNNs); popular methods include gated recurrent units (GRUs) and Long Short-Term Memory (LSTM) networks. The method developed in this paper can be viewed as an infinite-layer or continuum limit of a particular type of RNN; we describe this relationship next.

\subsection{Deriving non-autonomous Equation Discovery from RNNs}
Recurrent Neural Networks (RNN) were first introduced by \cite{Rumelhart:1986we} and have been used for processing sequential data. For input data, 
$x_t \in \mathbb{R}^n$ and a hidden state vector 
$h_t \in \mathbb{R}^m$ (typically initialized with $h_0 = 0$), a traditional sequence-to-label RNN is given by a discrete-time map
\begin{equation}\label{RNNm2o}
h_{t} = \phi(Wh_{t-1}+Ux_{t} + b) \text{ for } t \in [T],
\end{equation}
together with an output layer $\hat{y} = \sigma(Ah_{T} + b)$.
Here $\phi$ is a user-specified activation function, and for classification problems, $\sigma$ is typically the softmax function.
To train an RNN, we learn parameters $W$, $U$ and $b$. Typically RNNs are difficult to train due to a loss of long-term memory and suffer from computational issues in backpropagation through time, called \emph{exploding} or \emph{vanishing} gradients \cite{pmlr-v28-pascanu13}. 
Gated RNNs, such as the Long Short Term Memory network (LSTM) were developed in \cite{HochSchm97} to overcome the challenge of long term dependencies.  These more complex RNN architectures can be represented abstractly as a discrete-time map with parameter vector $\theta$---see \cite[Eq. (10.5)]{GoodBengCour16}:
\begin{equation}\label{RNNiter}
h_t = f( h_{t-1}, x_t;\theta) \text{ for } t \in [T].
\end{equation}
From \eqref{RNNiter}, we derive a continuous-time model as follows.  We first insert $N-1$ hidden layers between $h_{t-1}$ and $h_{t}$ and consider the discrete-time map
\begin{equation} \label{RNNiterdeep}
h_{t} = f( h_{ t - 1/N }, x_{t}; \theta) \text{ for } 
t = i/N \text{ with } i \in [N T].
\end{equation}
When $N = 1$, we recover \eqref{RNNiter}.  For $N \gg 1$, the model has a deep hidden-to-hidden transition \cite{Pascanu2014}; $N$ layers must be traversed to go from $h_{t-1}$ to $h_t$.  Next, we take a near-identity or residual network form of the right-hand side function $f$.  Essentially, $f(h,x;\theta) = h + \delta t \Phi(h,x;\theta)$ for a non-autonomous vector field $\Phi$ and $\delta t = N^{-1}$:
\begin{equation} \label{RNNdeepni}
h_{t} = h_{t-1/N} + N^{-1} \Phi( h_{ t - 1/N }, x_{t}; \theta) \text{ for } 
t = i/N \text{ with } i \in [N T].
\end{equation}
Finally, we take the \emph{infinite-layer} limit $N \to \infty$ and obtain the central equation in the \emph{non-autonomous equation discovery (NAED) method}:
\begin{equation}
\label{eqn:NAEDprop}
\frac{d}{dt} h(t)  = \Phi \left(  h(t),  x(t) ;\theta \right) \text{ for } t \in [0, T].
\end{equation}
We view the input signal $x(t)$ as a forcing term in a non-autonomous dynamical system governing a hidden variable $h \colon [0,T] \to \mathbb R^m$.  We introduce a function $\pi \colon \mathbb R^m \to  \mathbb R^{|\mathcal Y|}$ to assign a class label to the hidden variable evaluated at the final time, $\hat{y} =  \pi \left( h(T) \right)$.  The objective is to learn the right-hand side $\Phi$, parameterized by $\theta$, and the function $\pi$, so that given a new time signal $x(t)$, $t \in [0,T]$ we can estimate its class label, $y$.   In the NAED method, we represent the right-hand side function $\Phi$ using a predetermined \emph{dictionary}, a set of candidate functions, which is sufficiently large to capture a wide class of dynamics. There are a variety of choices for dictionaries; here we employ polynomial and Fourier basis functions.

In Section~\ref{sec2}, we describe the NAED method in more detail, including efficient gradient computation using the adjoint method.  In practice, we are given a time series, which we think of as a discretized time signal and we must also discretize the dynamical system to obtain a discrete-time approximation of the hidden variable. In this paper, we employ the optimize-then-discretize approach, where the gradient is computed analytically (see Theorem~\ref{p:grad}) using the continuous-time hidden variable and input time signal, and then evaluated using the time series and discretized hidden variable.  This in contrast to a discretize-then-optimize approach that begins with discrete-time models such as (\ref{RNNiter}) or (\ref{RNNiterdeep}), and then optimizes using gradients computed via backpropagation-through-time.

\subsection{Motivation and related work}
We motivate the NAED method in three ways:
\begin{enumerate}
\item As shown above, the NAED method can be viewed as an infinite-layer limit of a particular class of residual RNN architectures.  In the context of deep feedforward networks, this residual structure has been shown to improve robustness to noise and generalizability \cite{Haber}.  When we train the NAED method, we avoid common issues with training deep RNNs, such as vanishing/exploding gradients.  This allows NAED training to proceed without truncated backpropagation through time, gradient clipping, or other  heuristics.
\item While deep RNNs have achieved high accuracy rates on certain time series classification tasks, they are often difficult to interpret.  We seek to blend the high accuracy of deep RNN architectures with the interpretability of continuous-time dynamical system methods.  In particular, we illustrate in Section \ref{sec3} that trained NAED models can be interpreted graphically using phase portraits. Additionally, the differential equation form of the model enables us to prove that the outputs of the classifier are stable with respect to both deterministic and random perturbations. 
\item The dictionary representation of the vector field $\Phi$ is motivated by the literature on equation discovery.  The problem formulation and goal in equation discovery differs from ours; there one assumes that the data consists of observations of the state vector $h(t)$ of a continuous-time dynamical system.  Using this data, the goal is to learn the vector field $\Phi$.  This is a nonparametric regression problem, equivalent to finding a system of ordinary differential equations that fit the observations $h(t)$.  The Sparse Identification of Nonlinear Dynamics (SINDy) method assumes that $\Phi$ can be represented as a sparse linear combination of elements from a dictionary $\Xi$ \cite{EqnDiscovery}. Representing $\Phi$ with a dictionary requires fewer parameters than with a neural network.  In SINDy, training proceeds via an iteratively thresholded least squares method whose convergence has been established \cite{SINDY}.  In our Algorithm \ref{alg:sparsity}, we retain the iterative thresholding step from SINDy.  However, we generalize SINDy in the following way: we do not assume access to $h(t)$ at all, but rather the forcing function $x(t)$.  Learning $\Phi$ is a byproduct of our method, but the goal is to train a model whose predictions $\hat{y}$ match the true labels $y$.
\end{enumerate}

Continuous-time RNNs were proposed by Hopfield \cite{Hopfield1984} and studied by many authors---see \cite{Funahashi1993,Beer95} and references therein.  Early continuous-time RNNs were proposed as models of associative memory and hence are not directly comparable to the classifiers studied here.  Still, early continuous-time RNNs share two features with NAED: the models are expressed as systems of nonlinear differential equations, and inputs are treated as non-autonomous forcing terms.  Compared to NAED, early continuous-time RNNs have a rigid right-hand side structure that guarantees Lyapunov stability of the unforced system \cite{Hopfield1984}. In contrast, NAED has a flexible right-hand side $\Phi$ that we can often represent as a sparse linear combination of dictionary functions.

More recently, there has been a growing literature that connects deep and recurrent neural networks with ordinary differential equations (ODEs).  One branch of this literature seeks to apply ideas from dynamical systems theory to determine stable feedforward architectures \cite{Haber}, RNNs that do not exhibit chaotic dynamics \cite{CFN}, and RNNs that are constrained to be linearly stable \cite{antisym}.  The RNNs considered in these works \cite{CFN,antisym} do not involve ODEs.

Another branch stems from Neural ODEs or ODE-Nets  \cite{ChenRBD18}.  We view both NAED and Neural ODEs as infinite-layer limits of deep networks that are trained via the adjoint method rather than backpropagation.  Note that in Neural ODEs, the vector field is typically modeled using a (static) feedforward neural network (rather than with a dictionary), and the input is used as an initial condition (rather than a forcing term) to the ODE system. Recent efforts have sought to make Neural ODE techniques more practical for large-scale problems \cite{Oberman2020,Namboodiri2020,SNODE2020} and also to better understand the learning of genuinely continuous-time dynamics \cite{Tiemann2020}; we may be able to apply similar ideas to NAED in future work.

Recently, there has been some effort to generalize Neural ODE models to the RNN context.  \cite{Duvenaud2019} combines ODE-Net and RNN layers, instead of purely relying on ODEs as in NAED.  We also find continuous-time versions of GRU and LSTM models \cite{Moreau2019,Park2019,Pearlmutter2020}.  Compared with NAED, these architectures have more constraints on the right-hand side vector field $\Phi$.  Finally, the NAED dynamical system \eqref{eqn:NAEDprop} can be viewed as a special case of the recently proposed neural controlled differential equation (NCDE) model \cite{Lyons2020}.  Compared with NAED, the controlled differential equation allows for more general dependency of the hidden state $h(t)$ on the input $x(t)$.  While NCDE uses a neural network model of the vector field, NAED uses a dictionary.

Let us briefly outline the present paper.  In Section \ref{sec2}, we formally define the NAED method, establish existence/uniqueness of the method's solutions, compute gradients via the adjoint method, and also quantify the method's stability.  In Section \ref{sec3}, we report the results of several computational experiments that demonstrate the competitive performance of  NAED  with respect to widely used and/or related methods for time series classification.  We carry out these experiments both for synthetic data and for real data from the UCR Time Series Classification Archive \cite{UCR}. In these experiments,  NAED achieves similar or better accuracy than recurrent neural network methods (including LSTM and CFN architectures) and neural controlled differential equations (NCDE), with orders of magnitude fewer parameters. Additionally, we give examples of how  NAED  yields graphical interpretability in the form of phase portraits.  We conclude in Section \ref{sec4} with a discussion of the NAED method and ideas for future directions. Appendix~\ref{s:App} contains proofs of the Theorems given in this paper.

%%%%%%%%%%%%%%%%%%%%%%%%%%%%%%%%%%%%%%%%
\section{Non-autonomous equation discovery (NAED) method}\label{sec2} 
In this section, 
we describe our proposed non-autonomous equation discovery (NAED) method for time series classification, 
a gradient-based method for training it, our choice of dictionary in the NAED method, stability of the classifier, and a sparse version of the method. 

\subsection{NAED model for time signal classification}
We assume that we are given data of the form 
$\{x_i,T_i,y_i\}_{i\in [N]}$, where $x_i\colon [0,T_i]\rightarrow \mathbb{R}^n$ is a time signal and  $y_i \in \mathbb R^{|\mathcal Y|}$ is a probability mass function over the classes. In practice, $y_i$ will be a unit vector and $\argmax_i y_i$ will be the class or label.  Note that we allow for the possibility that the time signals have different lengths. We consider  the following non-autonomous dynamical system: 
\begin{subequations} \label{e:model}
\begin{align}
\frac{d}{dt} h_i(t) &= \Phi\left(h_i(t),x_i(t); \theta \right),  \qquad t \in [0,T] \\
\label{e:modelb}
h_i(0) &= h_{0}. 
\end{align}
For each time signal $x_i(t)$, we interpret the solution to \eqref{e:model}, $h_i(t) \in \mathbb R^m$ $\forall t \in [0,T_i]$ as a time-dependent hidden variable that is being \emph{forced} by the function $x_i(t)$.  The solution at time $T_i$ is used to make a class prediction $\hat y_i$ via
\begin{equation}
\label{e:softmax}
\hat y_i = \sigma \left( A h_i (T_i ) +b \right),
\end{equation}
\end{subequations}
where  
$A \in \mathbb R^{|\mathcal Y| \times m} $,   
$b \in  \mathbb R^{|\mathcal Y|}$, and 
$\sigma\colon \mathbb{R}^{|\mathcal Y|} \rightarrow \mathbb{R}^{|\mathcal Y|} $ is the softmax function, defined by
$[\sigma(x)]_i  = \frac{e^{x_i}}{\sum_{j}e^{x_j} }$.

We parameterize the vector field 
$\Phi \colon\mathbb R^m \times \mathbb R^n \to \mathbb R^m$ using a dictionary $\mathcal D = \{ \xi_j \}_{j \in [d]}$, with $\xi_j \colon \mathbb R^m \to \mathbb R$. 
We discuss specific choices for the dictionary, $\mathcal D$, in Section \ref{s:Dict}, but we have in mind, \eg, multivariate polynomials.  Let  $\theta = (\beta, B)$.  Composing the dictionary elements in a dictionary, $\Xi(h)= (\xi_1(h), \xi_2(h), \cdots , \xi_d(h)) \in \mathbb{R}^d$, we write
\begin{equation}\label{RHS}
\Phi( h,x; \theta ) = \beta \Xi(h)  + Bx, 
\end{equation}
where
$ \beta \in \mathbb R^{m \times d}$ 
and 
$B \in \mathbb R^{m \times n}$ 
are unknown coefficients.

To train the classifier, we must learn $\theta = (\beta, B)$, determining $\Phi$ via \eqref{RHS}, together with the parameters $A$ and $b$ in \eqref{e:softmax}.  We frame this learning problem as one of minimizing the following cross-entropy loss between labels $y_i$ and predictions $\hat{y}_i$:
\begin{equation} \label{e:Loss}
J(\Theta) =- \frac{1}{N}  \sum_{i\in [N]} \sum_{j \in \mathcal Y}  [y_i]_j \log[\hat{y}_i]_j =  - \frac{1}{N}  \sum_{i\in [N]} \sum_{j \in \mathcal Y}  [y_i]_j \log  [\sigma \left( A h_i (T_i ) +b \right)]_j.
\end{equation}
Here, $\Theta = \{ \beta, B, A, b\}$ represents all parameters to be learned.  It is understood that $h_i$ satisfies \eqref{e:model} for the forcing $x_i(t)$, $t \in [0, T_i]$. 

An important consideration is whether there exists a solution of the dynamical system in \eqref{e:model} with right-hand side given by \eqref{RHS}.  The following theorem gives a sufficient condition for the existence and uniqueness of a solution. 
\begin{thm}\label{prop:exun}
Assume $x \colon [0,t] \to \mathbb R^n$ is a continuous function. 
Let $K \subset \mathbb R^m$ be a compact set containing the initial point $h_0$
such that 
$\xi_i\colon \mathbb R^m \to \mathbb R$ is a locally Lipschitz continuous function on $K$ with Lipschitz constant $\mathcal L$ for every $i \in [d]$, \ie,  $\forall h_1, h_2 \in K$, 
$
| \xi_i(h_1) - \xi_i(h_2) | 
\leq \mathcal L \|h_1 - h_2\|$. 
Then there is an $\varepsilon>0$ such that the initial value problem in \eqref{e:model} has a unique solution defined on the interval $[-\varepsilon,\varepsilon]$.
\end{thm}
A proof of Theorem~\ref{prop:exun} is provided in Appendix~\ref{s:App}.

\subsection{Gradient computation and the adjoint method}
For the NAED time signal classifier, training can be formulated  as the ODE-constrained optimization problem,
\begin{equation} \label{e:Opt}
\min_{ \Theta = \{\beta,B,A,b\}} \quad J(\Theta), 
\end{equation}
subject to \eqref{e:model} where the objective function $J(\Theta)$ is defined in \eqref{e:Loss}. 
To employ a gradient-based optimization method, we need to compute $\nabla_\Theta J$. However, directly computing the gradient of $J$ with respect to $\Theta$ is complicated and computationally expensive because $J$ involves $h_i(T;\Theta)$, the solution to \eqref{e:model} at time $t=T$. 
An alternative method to compute $\nabla_\Theta J$ is to use the adjoint method, as we do in the following theorem.

\begin{thm} \label{p:grad}
The gradients of the objective function in \eqref{e:Loss} with respect to the unknown parameters:
$\beta \in \mathbb R^{m\times d}$, 
$B \in \mathbb R^{m\times n}$, 
$A\in \mathbb R^{|\mathcal Y| \times m}$, and 
$b \in  \mathbb R^{|\mathcal Y|}$
are given by
\begin{subequations} \label{e:grad}
\begin{align}
\nabla_\beta J &= -\sum_{i \in [N]} \int_0^{T_i} \lambda_i(t)\Xi(h_i(t))^t   \ dt \\
\nabla_B J  &= - \sum_{i \in [N]} \int_0^{T_i} \lambda_i(t) x_i(t)^t   \ dt 
\end{align}
\begin{align}
\nabla_A J  &= \partial_A J =  - \frac{1}{N}  \sum_{i\in [N]}\left( y_i -\sigma(Ah_i(T_i) +b)\right) h_i(T_i)^t \\
\nabla_b J  &= \partial_b J  =  - \frac{1}{N}  \sum_{i\in [N]} (y_i -\sigma(Ah_i(T_i) +b))
\end{align}
\end{subequations}
where $\lambda_i(t)$ for  $t\in [0,T_i]$ is a solution to the adjoint equation, 
\begin{subequations} \label{e:adj}
\begin{align} 
\frac{d}{dt} \lambda_i(t) &= -[\beta D_h\Xi(h) ]^t\lambda_i(t) \\
\lambda_i(T_i) &=  -\frac{1}{N} A^t (y_i-\sigma(Ah_i(T_i)+b)).
\end{align}
\end{subequations}
\end{thm}
A proof of Theorem~\ref{p:grad} is provided in Appendix~\ref{s:App}. The gradients from Theorem~\ref{p:grad} are used with an optimization method to minimize the cross-entropy loss function \eqref{e:softmax} and thereby train the model.

\subsection{Dictionary choice} \label{s:Dict}
In the NAED model, the right-hand side of the dynamical system is given by 
$\Phi( h,x; \theta) = \beta \Xi(h)  + Bx$; see \eqref{RHS}. The first term is a linear combination of dictionary functions, 
$\Xi(h)= (\xi_1(h), \xi_2(h), \cdots , \xi_d(h)) \in \mathbb{R}^d$. 
There is tremendous freedom in selecting the dictionary functions and this choice is paramount to the model. We tested NAED using two different dictionaries, a \emph{polynomial dictionary} and a \emph{Fourier dictionary}, described now in turn.

The polynomial dictionary consists of all possible polynomials of $h \in \mathbb{R}^m$ up to  $k$-th order. For $h\in \mathbb R^m$, the dictionary is 
$\Xi(h) = [1,P_1(h),P_2(h),\dots P_k(h)]$,
where $P_k(h)$ is a basis for homogeneous polynomials of degree $k$. We choose the basis $P_k(h)$ to consist of the  $\binom{k+m-1}{m-1}$ basis elements of the form
$\frac{1}{\alpha_1! \cdots \alpha_m! } h_1^{\alpha_1} \cdots h_m^{\alpha_m}$, 
where
$\sum_{i=1}^m \alpha_i = k$, 
as appearing in Taylor's theorem. 
For instance, if $m=2$,  $P_2(h)$ refers to the quadratic polynomials
$P_2(h) = \left[h_1^2/2~,~ h_1h_2~,~ h_2^2/2\right]$.

Alternatively, we can consider a Fourier dictionary. Using separation of variables for the function 
$\Phi\colon \mathbb{R}^m\rightarrow \mathbb{R}$, we write
$\Phi(h) = f_1(h_1)f_2(h_2)\cdots f_m(h_m)$,
where $f_i\colon \mathbb{R}\rightarrow \mathbb{R}$ for $i = 1,\ldots, m$. We approximate $f_i(x)$ by a finite linear combination of Fourier basis functions, 
$
f_i(x) = a^i_0 + \sum_{k = 1}^K  a^i_k\cos\left(2\pi k x/ L\right) + b^i_k\sin\left( 2\pi k x/L\right),
$
where $L$ is the period of $f_i(x)$. Each row of the vector $\beta \Xi(h)$ appearing in the RHS of the dynamical system \eqref{e:model} can be written as 
$\prod_{i=1}^m f_i(h_i)$,
% $$ \prod_{i=1}^m~ \left(a^i_0 + \sum_{k = 1}^K  a^i_k\cos\left( 2\pi k h_i/L\right) + b^i_k\sin\left( 2\pi k h_i/L\right) \right),
% $$
where the coefficients $a^i_0$, $a^i_k$, and $b^i_k$ correspond to entries of $\beta$. 
In other words, our dictionary $\mathcal{D}$ consists of functions given by the outer product of harmonic functions, 
\begin{equation*} \footnotesize
\Xi(h) =
\begin{pmatrix}
\sin\Big(2\pi k_1 h_1/L\Big) \\
\cos\Big(2\pi k_1 h_1/L\Big)
\end{pmatrix}
\otimes
\begin{pmatrix}
\sin\Big(2\pi k_2 h_2/L\Big) \\
\cos\Big(2\pi k_2 h_2/L\Big)
\end{pmatrix}
\otimes
\cdots 
\otimes 
\begin{pmatrix}
\sin\Big(2\pi k_m h_m/L\Big) \\
\cos\Big(2\pi k_m h_m/L\Big)
\end{pmatrix},
\end{equation*}
where $k_i \in [K]$.
For instance, for $m=2$, and $K= 1$, the dictionary consists of the following 9 functions: 
{\scriptsize
\begin{align*} 
\Xi(h) &=  \Big[  1, 
\cos\left( 2\pi h_1/L\right),  \sin\left( 2\pi  h_1/L\right), 
\cos\left(2\pi h_2/L\right), 
\sin\left( 2\pi  h_2/L\right), 
\cos\left( 2\pi h_1/L\right)
\cos\left( 2\pi h_2/L\right),\\
&\sin\left( 2\pi h_1/L\right) 
 \cos\left( 2\pi h_2/L\right), 
\cos\left(2\pi h_1/L\right) 
\sin\left( 2\pi  h_2/L\right), 
\sin\left( 2\pi h_1/L\right)
\sin\left( 2\pi  h_2/L\right) \Big]. 
\end{align*}}

Note that by the Stone-Weierstrass theorem, the polynomial dictionary and Fourier dictionary are dense in the space of continuous functions and $L^2$, in the limiting case where $k\to \infty$ and $K \to \infty$, respectively.  By choosing these parameters sufficiently large, all smooth dynamical systems can be represented as accurately as is needed.

We would like to apply Theorem~\ref{prop:exun} to guarantee the existence of a unique solution to \eqref{e:model}. 
Assuming that the time signal $x$ is continuous, it is enough to choose a dictionary that satisfies the Lipschitz continuity assumption. 
If we use the Fourier dictionary, then the Lipschitz constant is approximately $\mathcal{L}\approx  \frac{2\pi K}{L}$. 
In this case, our model \eqref{e:model} has a unique solution until time $T$, provided we  initialize $\beta$ with sufficiently small values. 
On the other hand, if we use the Polynomial dictionary, there are two cases. If only linear terms are used in the dictionary, then the right-hand side is Lipschitz continuous and a unique solution exists on the time interval $[0,T]$. However, if we use higher-order polynomials in the dictionary, the right-hand side is only locally Lipschitz and Theorem~\ref{prop:exun} can only guarantee a solution on a short time interval; the solution may blow up in finite time.  In the numerical results in Section \ref{sec3}, we will observe that models with the Fourier dictionary are generally more accurate and less sensitive to initialization than models with a nonlinear Polynomial dictionary. 

\subsection{Stability of the NAED method} \label{s:stability}
Dynamical systems theory can be used to prove that a given NAED classifier $x \xmapsto{\mathscr{C}} y$ is stable to noise; below we do this for both deterministic and stochastic perturbations.
For $p \in [1,\infty)$, let $L^p\left([0,T]; \mathbb R^n \right)$ denote the Bochner space of continuous $\mathbb R^n$-valued  functions with norm  
$\|x\|_{L^p\left([0,T]; \mathbb R^n \right)} := \left(\int_0^T |x(t)|^p dt \right)^\frac{1}{p}$.

\begin{thm}
\label{p:stability}
Consider a NAED classifier $\mathscr{C}\colon L^1\left([0,T]; \mathbb R^n\right) \to \mathbb R^{|\mathcal Y|}$ with dictionary $\Xi\colon \mathbb R^m \to \mathbb R^d$ that is Lipschitz continuous with constant $\mathcal L$. 
The classifier $\mathscr{C}$ is Lipschitz continuous with constant $L>0$ defined in the proof. That is, if we have a time signal $x(t)$ and a noise corrupted version, $\tilde x(t) = x(t) + \eta(t)$, then
$ | \mathscr{C}(\tilde x) - \mathscr{C}(x)  | \leq L  \| \eta \|_{L^1\left([0,T]; \mathbb R^n \right)}$.
\end{thm}

\begin{thm} 
\label{p:stabilityWiener}
Consider a NAED classifier $\mathscr{C}\colon L^1\left([0,T]; \mathbb R^n\right) \to \mathbb R^{|\mathcal Y|}$ with dictionary $\Xi\colon \mathbb R^m \to \mathbb R^d$ that is Lipschitz continuous with constant $\mathcal L$. 
%Consider a NAED classifier $\mathscr{C}$ with dictionary $\Xi\colon \mathbb R^m \to \mathbb R^d$ that is Lipschitz continuous with constant $\mathcal L$. The classification mapping $x \xmapsto{\mathscr{C}} y$ is Lipschitz continuous on the space $L^1\left([0,T]; \mathbb R^n\right)$ with constant $L>0$ defined in the proof. 
Let $W_t$ denote the Wiener process in $\mathbb{R}^d$.  Consider a time signal $x(t)$ and a version corrupted by Gaussian white noise, $\tilde x(t) = x(t) + \eta(t)$, where $\eta(t) dt = dW_t$.  Then 
$| \mathscr{C}(\tilde x) - \mathscr{C}(x)  | \leq L  \sup_{0 \leq s \leq T} |W_s|$ and 
 $P \left( | \mathscr{C}(\tilde x)  - \mathscr{C}(x) | \geq r 
 \right) \leq 2 d e^{-r^2 / 2 d T L^2}$, with constant $L > 0$ defined in the proof.
\end{thm}
Proofs of Theorem~\ref{p:stability} and \ref{p:stabilityWiener} are provided in Appendix~\ref{s:App}. Theorems~\ref{p:stability} and \ref{p:stabilityWiener} can be further interpreted in terms of classification stability  as  follows. 
Suppose that for a given time series, $x$, the NAED classifier gives the  estimate $\hat y = \mathscr{C}(x)$ (a probability vector). Further, suppose that $\max_i \mathscr{C}(x)_i$ is uniquely attained so that the distance between $\mathscr{C}(x)$ and the decision boundary is positive. Then there exists a positive constant, $\varepsilon >0$, such that for any corrupted time signal $\tilde x(t) =  x(t) + \eta(t)$ with $\| \eta\| \leq \varepsilon$ the two estimates $\mathscr{C}(x)$ and  $\mathscr{C}(\tilde x)$ have the same maximum component, and so the assigned class does not change for the corrupted time signal. Here, the corruption can be either deterministic (Theorem~\ref{p:stability}) or stochastic (Theorem~\ref{p:stabilityWiener}).

\subsection{Sparse NAED method} \label{s:sparsity}
The main task in our proposed learning method is to find the right-hand side (rhs) of the underlying non-autonomous dynamical system in \eqref{RHS}, where the rhs is assumed to be a linear combination of dictionary terms. 
Here we explore the idea of imposing sparsity on the dictionary coefficients, with the goal of finding a \emph{simple representation} of the underlying dynamics. As in equation discovery methods, we are motivated by the observation that most equations describing physical phenomena involve only a few relevant terms so that the rhs is sparse in the set of all possible functions. Imposing this assumption, we learn a model that balances accuracy and parsimony. Additionally, the sparsity assumption on the dictionary coefficients helps to  prevent overfitting on the training dataset, leading to a method that is more robust to noise. Moreover, by assuming sparsity, we also obtain more interpretable dynamical models. 

To develop a practical method to promote sparsity in the dictionary coefficients, we adopt the idea of iterative thresholding from \cite{EqnDiscovery,SINDY}. 
The resulting algorithm is given in Algorithm \ref{alg:sparsity}. In Algorithm \ref{alg:sparsity}, entries of $\beta$ with magnitude less than $\lambda>0$ are thresholded to zero. This procedure is repeated until $\beta$ has converged.  In general, increasing $\lambda$ trades  accuracy for sparsity.  The optimal value of $\lambda$ will thus depend on the problem and data at hand.  In practice, we use cross-validation to tune the value of $\lambda$.  We note that the convergence of the algorithm depends on the value of $\lambda$. 

\begin{algorithm}[t]
\caption{Sparse NAED method for time signal classification}\label{alg:sparsity}
\begin{algorithmic}[]
\State {\bf Input:} cut-off value, $\lambda >0$ and initial parameters, $\Theta = \{\beta, B, A, b\}$.
\While{not converged:}

\State {\bf (descent step)} Take a step for the parameters $\Theta$ according to a chosen method to minimize $J(\Theta)$ in \eqref{e:Loss}. {\it E.g.}, for the gradient descent method with stepsize $t>0$, update the parameters according to  
$\Theta \leftarrow \Theta - t \ \nabla_\Theta J$.

\State {\bf (threshold step)} We threshold the $\beta$ parameter values by setting
$$
\beta_{ij} \leftarrow
\begin{cases}
\beta_{ij}, & \textrm{if } |\beta_{ij}| \geq \lambda \\
0, & \textrm{if } |\beta_{ij}| < \lambda 
\end{cases}.
$$
\EndWhile
\end{algorithmic}
\end{algorithm}

%----------------------------------------
\section{Computational experiments}\label{sec3}
In this section, we demonstrate our proposed NAED method  on a variety of datasets: 
synthetic datasets derived from dynamical systems and partial differential equations (Section \ref{s:SynData}), 
a noisy synthetic dataset derived from a dynamical system (Section \ref{data:noise}), and
UCR Archive Datasets (Section \ref{data:UCR}). 
We demonstrate that our method is interpretable and attains results with accuracy comparable to or better than the RNN, LSTM, CFN and NCDE methods, using substantially fewer parameters. 
Next we describe details of our implementation; also note that our source code is available at \verb+https://github.com/rkyoon12/NAED-Method+.

\subsection{Implementation details} \label{s:AlgFrame}
We implemented the NAED method, described in Section \ref{sec2}, using TensorFlow. To solve the optimization problem, we used the ADAM optimizer with gradient computed as in Theorem~\ref{p:grad}. The gradient computation requires us to solve both the \emph{forward ODE} \eqref{e:model} for $h\colon[0,T] \rightarrow \mathbb R^m$  and  the \emph{adjoint ODE} \eqref{e:adj} for $\lambda\colon[0,T] \rightarrow \mathbb R^m$. To approximate the solution of the forward and adjoint ODEs, we used the fourth-order Runge–Kutta (RK4) method, implemented via \verb+tfs.integrate.odeint_fixed+  in the \verb+tensorflow_scientific+ library. To approximate $x(t)$ at times $t$ not in the sampled time series data, we use linear interpolation
$x(t) \approx \frac{ (t_{n+1}-t)x_n +  (t-t_{n})x_{n+1}}{t_{n+1}-t_{n}},$
$\quad t_n\leq t \leq t_{n+1}. $
In all numerical examples we fixed the initial condition for the hidden state in \eqref{e:modelb} to be $h_0 = 0$.
For the ADAM method, we used a learning rate $\in \{0.001,0.005,0.01,0.05,0.1,0.5\}$.

For each dataset, we train the NAED model several times for different dimensions, $m$, of the hidden state, largest degree of polynomials $k$, or the maximum number of Fourier basis terms $K$.  We report the results for several such models. 

\paragraph{Initialization.} 
As described at the end of Section \ref{s:Dict}, the choice of dictionary functions and coefficients has a significant effect on the convergence of the method.  In particular, large values of $\beta$ can cause the solution of the forward ODE \eqref{e:model} to blow up in finite time. The time duration $\varepsilon$, as guaranteed by Theorem~\ref{prop:exun} for a bounded solution, is inversely proportional to the norm of $\beta$ and $B$. Hence we initialize parameters to be small to guarantee a bounded solution  to \eqref{e:model} until the final time $T$. 
Let $\mathcal{U}[r,s]$ denote the uniform distribution on the interval $[r,s]$.  For either the linear polynomial dictionary  or the Fourier basis dictionary, the Lipschitz constant for each dictionary function is approximately $ \mathcal{L}\approx 1$, so we initialize the parameters $\beta, B, A \stackrel{iid}{\sim} \mathcal{U}[-1,1]$ and $b \stackrel{iid}{\sim} \mathcal{U}[0,1]$. When the dictionary involves higher-degree polynomials, we initialize $\beta, B \stackrel{iid}{\sim} \mathcal{U}[-0.1,0.1]$ and $A, b \stackrel{iid}{\sim} \mathcal{U}[-1,1]$.

\paragraph{Other methods.} 
We implemented the RNN and LSTM methods in TensorFlow, using 
\verb+tf.keras.sequential+ 
with \verb+keras.layers.rnn+ 
and \verb+keras.layers.LSTM+ layers.  Using the description in \cite{CFN}, we developed our own implementation of the CFN method in TensorFlow. We train the NCDE method using the published code \cite{Lyons2020}; we implemented this in PyTorch using the \verb+torchcde+ library.  
We trained each model using the cross-entropy loss function, the ADAM optimization method, and the default initialization. The models were trained until convergence of the loss function.  For all considered datasets and all methods, we report the best result among our numerical experiments after varying hyper parameters such as network depth and width.

\subsection{Synthetic datasets} \label{s:SynData}
\subsubsection{ Forced harmonic  oscillator  }\label{data:OSC}
We consider a forced oscillator with position $u(t)$ satisfying 
\begin{subequations} 
\label{e:ForcedOscillator}
\begin{align}
& \ddot u + \gamma \dot u + \omega^2 u = x(t)   \\
& u(0) = \dot u(0) = 0, 
\end{align}
\end{subequations}
where $\gamma$ is the damping coefficient, $\omega^2$ is the undamped angular frequency, and $x(t)$ is a specified forcing. To form the ground-truth labels, we record whether the position of $u(T)$ at the final time $t = T$ is positive or negative,
\begin{equation} \label{e:output}
y = \begin{cases}
(1,0) & u(T) > 0 \\
(0,1) & u(T) < 0 
\end{cases}. 
\end{equation}
With the above framework, we generate a synthetic dataset as follows. Fix $K \in \mathbb N$, 
$\gamma >0$, 
$\omega>0$, and
$T> 0$. 
For a forcing of the form 
$x(t) = \sum_{k=1}^K A_k \sin(\alpha_k t)$, $t\in [0,T]$, 
where $A_k $ are randomly chosen amplitudes and $\alpha_k$ are randomly chosen forcing frequencies, we numerically solve \eqref{e:ForcedOscillator} for $u(t), \ 
t \in (0,T]$ and compute $y$ via \eqref{e:output}. 
We choose  
$K=2$, 
$\gamma =0.2$, 
$\omega = 1$, 
$T=10$,
$A_k\stackrel{iid}{\sim} \mathcal{N}(0,1)$, and $\alpha_k \stackrel{iid}{\sim} \mathcal{N}(0,1)$.  In this paper, we use $\mathcal{N}(\mu,\sigma^2)$ to denote the normal distribution with mean $\mu$ and variance $\sigma^2$.
The process is repeated $N=10000$ 
times to create a dataset with $8000$ training examples and $2000$ test examples. 
\begin{table}[t!]
\center
\begin{tabular}{|c|c|c|c|}
\hline
Methods&Train&Test& $\#$ params\\
\hline
\textbf{EqnDis Poly (2,1) }& \textbf{0.9994} & \textbf{0.9905} & 14\\
EqnDis Poly (3,1) & 0.9831 & 0.9585 & 23\\
EqnDis Poly (4,1) & 0.9800 & 0.9040 & 34\\
EqnDis Poly (2,2) & 0.9817 & 0.9700 & 20\\
\hline
EqnDis Fourier (2,1) & 0.9614 & 0.9670& 26\\
EqnDis Fourier (2,2) &0.9365	& 0.9345&58\\
\hline
 RNN (1,5)           & 0.9741 & 0.9715 & 41\\
 LSTM (1,5)         & 0.9791 & 0.9750 &146 \\
 CFN (7,5)           & 0.9113 & 0.9180 & 891 \\
 NCDE (32,32-1)       &  0.9919 & 0.9854 & 1221 \\      
 \hline
 \end{tabular}
 \caption{A comparison of the accuracy (training and test datasets) and number of parameters for various methods on the synthetic dataset based on the forced harmonic oscillator. In this and in subsequent tables, we use boldface to indicate methods with the highest accuracy.  See Section \ref{data:OSC}.}
 \label{table:OSC}
 \end{table}

In Table \ref{table:OSC}, we tabulate the accuracy and number of trained parameters for various methods on this dataset.  Note that the total number of parameters for the NAED method is given by
$$
\# \textrm{params} = 
\text{dim}(\beta) + \text{dim}(B) + \text{dim}(A) + \text{dim}(b) =
d\times m + n \times m + m \times |\mathcal Y| + |\mathcal Y|.
$$ 
In the first column of Table \ref{table:OSC}, additional information about each method is summarized. 
For the NAED Method with Polynomial dictionary, the parenthetical numbers are ($\# $ of units in hidden layer,  maximum degree of polynomial in dictionary). The first row block of Table \ref{table:OSC} is for the NAED method while varying either the dimension of the hidden units or the maximum degree of the polynomial entries. We observe that the Polynomial dictionary with a two-dimensional hidden state and polynomials up to degree one produces the best accuracy. This model also has the smallest number of parameters of all methods tested. This result might be expected as it agrees with the ground-truth model (harmonic oscillator). For the NAED method with the Fourier dictionary, the parenthetical numbers refer to ($\# $ of units in hidden layer, largest multiplier $K$) where the dictionary consists of Fourier terms with frequency $\omega= \frac{L}{K},\dots, L$. 
It is natural to choose $L=10$ because we handle the hidden state $h$ on the time interval $[0,10]$. 
For the RNN, LSTM, and CFN methods, the parenthetical numbers represent ($\#$ of hidden layers, $\#$ of units). For the NCDE method, the parenthetical numbers represent ($\#$ of units, \emph{width}-\emph{depth} of neural network for vector field). Note that repeated runs with different values of these hyperparameters were carried out, but we report only the hyperparameters for the models with the best test accuracy. We observe that all methods performed remarkably well for this simple dataset. 

\begin{figure}[t!]

\centering
\includegraphics[width=0.9\linewidth]{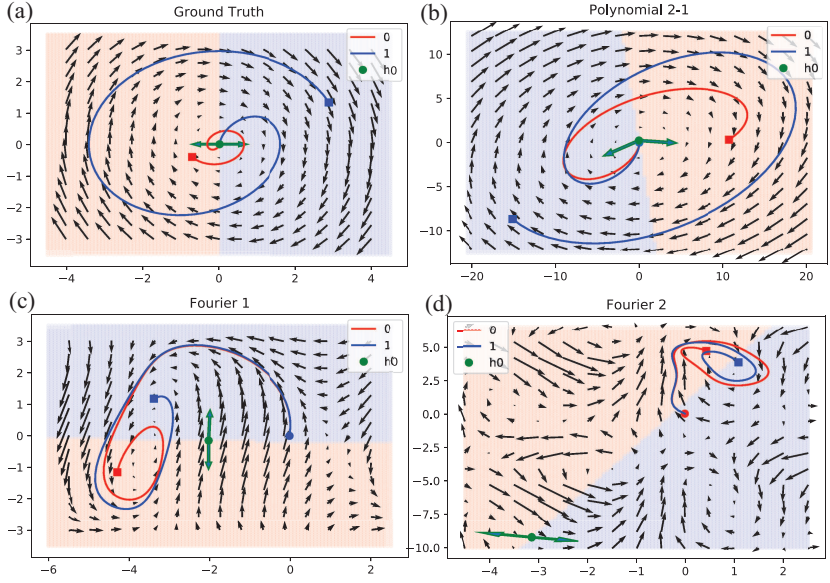}

\caption{In four subplots, labeled {\bf (a)}--{\bf (d)}, we plot the vector field $h \mapsto \beta \Xi(h)$  in \eqref{e:model} for different choices of dictionary $\Xi$.  In each plot,  two example solution trajectories are given (one for each class), vectors used for the decision are drawn, and the class regions are shaded in red and blue.   
{\bf (a)} Polynomial dictionary with ground truth initialization. 
{\bf (b)} Polynomial dictionary with random initialization. {\bf (c)} Fourier dictionary with $K=1$. 
{\bf (d)} Fourier dictionary with $K=2.$  See Section \ref{data:OSC} for details.}
\label{fig:OSC}
\end{figure}

We can visualize our model using phase portraits; examples are given in Figure \ref{fig:OSC}.  
Here, the black arrows represent the autonomous part of the learned vector field, $h \mapsto \beta \Xi(h)$.
Also plotted in color are solution trajectories. Note that all samples are initially at the origin and the final positions at $T=10$ are indicated by a square. The class associated with each sample is indicated in the legend. 
The classification decision is made using the
final state of a trajectory via the probability vector,  
$\tilde{y} = \sigma \left( A h (T ) +b \right)$.
Writing 
$A h (T) +b = A \left(   h(T) + A^{-1} b \right) = A \left(   h(T) - h_0  \right) $
where $h_0 = - A^{-1} b$, we see that the softmax function is being applied to the vector 
$ \begin{pmatrix} a_1^t (h(T) - h_0) \\ a_2^t (h(T) - h_0) \end{pmatrix}$, 
where $a_i$ is the $i$-th row of $A$. 
We can visualize this decision in  Figure \ref{fig:OSC} as follows. 
At the point $h_0$, we draw the two rows of $A$ as green vectors. These vectors partition $\mathbb R^2$ into two regions; (each representing a class); we shade the region representing class 0 in red and class 1 in blue.
In Figure \ref{fig:OSC}, we observe that the the final states of the chosen  trajectories belong to the correctly identified partition component. 

We now remark on the identifiability of our model. Recall that a statistical model is said to be \emph{identifiable} if the parameter values uniquely determine the probability distribution of the data. For an identifiable model, it is in principle possible to learn the ground-truth parameters used to construct the data.  Also recall that the goal of our algorithm is not to learn the mapping $h \to \beta \Xi(h)$, but rather the mapping $x \rightarrow y$. Since only the solution of the forward ODE at the final time is used to make this prediction, the learned vector field can differ from the ground truth vector field. 
If we consider Figure \ref{fig:OSC}{\bf (b)}, the learned vector field closely agrees with the ground truth vector field {\bf (a)}, up to conjugation by an orthogonal matrix.  The eigenvalues of $\beta \Xi(h)$ for in {\bf (a)} are $ \lambda = -0.1\pm i 0.995$ which are close to the eigenvalues of  $\beta \Xi(h)$ for {\bf (b)}, given by  $\lambda = -0.1015\pm i1.001$. 
However, the vector fields in {\bf (c)} and {\bf (d)} are seen to differ from {\bf (a)} considerably. 

\subsubsection{ Forced Van der Pol oscillator}\label{data:VDP}
Consider the forced Van der Pol oscillator with position $u(t)$ satisfying 
 \begin{subequations} 
\label{e:ForcedVanderPolr}
\begin{align}
& \ddot u - \mu (1-u^2)\dot u +  u = x(t)   \\
& u(0) = \dot u(0) = 0
\end{align}
\end{subequations}
where $\mu = 0.3$ controls the strength of nonlinear damping. We choose the forcing $x(t)$ as in Section \ref{data:OSC} and, at time $T = 10$,  we define the label $y$  as in \eqref{e:output}. 

\begin{table}[t!]\label{table:VDP}
\center
\begin{tabular}{|c|c|c|c|}
\hline
Method&Train&Test& $\#$ params\\
\hline
EqnDis Poly (2,1) & 0.9030  & 0.854  & 14\\
EqnDis Poly (3,1) &  0.9100 & 0.8675  & 23\\
EqnDis Poly (4,1) & 0.8790 & 0.882 & 34\\
EqnDis Poly (2,2) & 0.7458&0.765 & 20\\
EqnDis Poly (2,3) & 0.8237 & 0.8215& 28\\
\hline
EqnDis Fourier (2,1)&0.9045 & 0.8975 & 26\\
\textbf{EqnDis Fourier (2,2)} & \textbf{0.9830}&  \textbf{0.9860}&58\\

\hline
 RNN (5,7)           & 0.9729 & 0.9600 & 491\\
 LSTM (1,5)         & 0.9603 & 0.9495 &146 \\
 CFN (7,5)           & 0.9345& 0.9350 & 1,177\\
 NCDE (32, 32-1)    & 0.9821 & 0.9745 & 1,221\\
 
 \hline
 \end{tabular}
 \caption{A comparison of the accuracy and number of parameters for various methods on the forced Van der Pol synthetic dataset. See Section \ref{data:VDP}.}
 \end{table}
As shown in Table \ref{table:VDP}, the best accuracy for the forced Van der Pol dataset is obtained with the Fourier (2,2) dictionary. It is a remarkable result in that NAED only uses 58 parameters.  On the other hand, the second best result trains roughly 20 times more numbers of parameters.  Since the true system is nonlinear, it is not surprising to see strong performance from the Fourier dictionaries, which contain sums and products of trigonometric functions.  Due to the presence of nonlinear polynomials in the Van der Pol system, we might expect that the best dictionary would be the Polynomial (2,3) dictionary. However, as discussed in Section \ref{s:Dict}, the nonlinear entries in the dictionary cause the right-hand side of \eqref{e:model} to be only locally Lipschitz continuous, so that Theorem~\ref{prop:exun} can only guarantee a solution on a short time interval. Since the class prediction is made using  \eqref{e:softmax}, \ie, it depends on the hidden variable $h(t)$ at time $t=T$, premature blowup of solutions spoils the learning process. The first block in Table \ref{table:VDP} shows that linear Polynomial dictionaries beat nonlinear ones.  We obtained the best accuracy with a more complex Fourier dictionary; both Fourier dictionaries outperformed all polynomial dictionaries on this problem.

% The learned vector field using the corresponding dictionary is shown in Figure \ref{fig:VDP}.
% \begin{figure}[h!]\label{fig:VDP}
% \includegraphics[width=1\linewidth]{vdp.eps}
% \caption{{\it Show the vector field for RHS of \eqref{e:model} without forcing term, that is vector field of $\beta \Xi(h)$ trained on VDP dataset.  (a) $\beta$ is ground truth. (b) $\beta$ is learned with EqnDis Fourier(2-1). (c) $\beta$ is learned with EqnDis Fourier(2-2) }}
% \label{fig1}
% \end{figure}

%  in the variables $(u_1(t),u_2(t),u_3(t))$:

\subsubsection{ Forced Lorenz}\label{data:Lorenz}
Consider the forced nonlinear Lorenz system with positive parameters $(\sigma, \rho, \beta)$:
\begin{subequations} 
\label{e:ForcedLorenz}
\begin{align}
& \dot u_1  = \sigma(u_2-u_3) + x(t)\\
&\dot u_2 = u_1(\rho -u_3) -u_2\\
& \dot u_3 = u_1u_2 -\beta u_3\\
& u_1(0) = u_2(0)  = u_3(0) = 1
\end{align}
\end{subequations}
The first coordinate is forced by  
$x(t) =4 \sum_{k=1}^K A_k \sin(\alpha_k t)$, 
$t \in [0,T]$,
where $A_k \stackrel{iid}{\sim} \mathcal{N}(0,1)$ and $a_k \stackrel{iid}{\sim} \mathcal{N}(0,1)$. 
Using the position of $u_1(t)$ at the final time $T=10$, we define the label $y$ as in \eqref{e:output}. To generate the synthetic data, we choose parameters $\sigma = 5, \beta = 1.3$ and $\rho = 10$. Note that this dataset is not balanced in each class; the training data consist of $6348$ and $1652$ instances in classes $0$ and $1$, respectively, and the  test data contains $1566 $ and $434$ instances in classes $0$ and $1$, respectively.

\begin{table}[t!]
\center
\begin{tabular}{|c|c|c|c|}
\hline
Method&Train&Test& $\#$ params\\
\hline
EqnDis Poly (2,1) & 0.8388& 0.8365 & 14\\
EqnDis Poly (3,1) &  0.8252 & 0.8160  & 23\\
EqnDis Poly (4,1) & 0.8321 & 0.8215 & 34\\
EqnDis Poly (2,2) &  0.8522&0.847 & 20\\
EqnDis Poly (3,2) & 0.8546 & 0.8535 & 41\\
\hline
EqnDis Fourier (2,1) & 0.8861 & 0.8945 & 26\\
\textbf{EqnDis Fourier (3,1) }& \textbf{0.9051} & \textbf{0.9050}& 92\\
EqnDis Fourier (2,2) &0.8517 &0.8439 &58\\

\hline
 RNN (2,10)           & 0.7937 & 0.7799 & 341\\
 LSTM (1,10)         & 0.9306 & 0.9359  &491\\
 CFN (2,10)           & 0.8080& 0.7965 &781\\

 NCDE (16,16-1)     & \textbf{0.9434} & \textbf{0.9369} &595\\
 \hline
 \end{tabular}
\caption{A comparison of the accuracy and number of parameters for various methods on the synthetic dataset based on the forced Lorenz equation. See Section \ref{data:Lorenz}.}
\label{table:Lorenz}
 \end{table}
As shown in Table \ref{table:Lorenz}, the highest accuracy for different choices of dictionaries and parameters in the NAED method is obtained by Fourier (3,1). This result demonstrates that complex dictionary entries are required to capture the nonlinearity in the underlying dynamics. It is remarkable that the NAED methods produced comparable results to the other methods using far fewer parameters, although it does not exceeded the classification accuracy of the LSTM method.

\subsubsection{Forced Lotka-Volterra equations}\label{data:LV}
Consider the forced Lotka-Volterra system, 
\begin{subequations} 
\label{e:ForcedLV}
\begin{align}
& \dot u_1  = \alpha u_1-\beta x(t) u_1u_2  \\
&\dot u_2 = \delta x(t) u_1u_2 -\gamma u_2, 
\end{align}
\end{subequations}
with initial condition $(u_1(0), u_2(0)) = (5,4)$, 
$x(t) =\left( \sum_{k=1}^K A_k \sin(\alpha_k t)\right)^2\geq 0$, and parameters $(\alpha, \beta, \delta, \gamma) = (0.8,0.1,0.01,1.1)$. 
We sample $A_k ,a_k \stackrel{iid}{\sim} \mathcal{N}(0,0.5)$.  After numerically solving up to time $T=10$, we set the ground truth label  $y$ via the indicator function for  $\text{argmax} (u_1(T), u_2(T))$.  
Note that the $x(t)$ appears as a coefficient in the nonlinear terms.  If we introduce the additional variable $u_3(t) = x(t)$,  the forcing occurs linearly, 
\begin{align*}
& \dot u_1  = \alpha u_1- u_1u_2 u_3 & \quad u_1(0) &= 5\\
&\dot u_2 = u_1u_2u_3 -\gamma u_2 & \quad u_2(0) &= 4 \\
&\dot u_3 = \dot{x}(t) & \quad u_3(0) &= x(0)=0
\end{align*}
This system suggests that we consider $\dot{x}(t)$ as the time series input data. Hence we train the model using, in turn, either $x(t)$ or $\dot{x}(t)$ as the input. We generate $\dot{x}(t)$ using the derivative of $x(t)$ computed by hand.

 \begin{table}[t!]
 \center
\begin{tabular}{|c|c|c|c|}
\hline
Method&Train&Test& $\#$ params\\
\hline
EqnDis Poly (2,1) & 0.8835 & 0.8860  & 14\\
EqnDis Poly (2,3) &  0.8785& 0.8835& 28\\
EqnDis Fourier (2,1) &0.9600 &0.9539 &26\\
\textbf{EqnDis Fourier (3,1)} &\textbf{0.9805}&\textbf{0.9739} &92\\
\hline
 RNN (5,32)           & 0.8192 & 0.8045 & 9,441\\
 LSTM (1,47)         &0.9614&0.9595 &9,260\\
 CFN (5,20)           & 0.9295& 0.9184 &9,081\\
 NCDE (32,32-1)   & \textbf{0.9838} & \textbf{0.9789} & 1,221\\
 \hline
 \end{tabular}
 \vspace{10pt}

 \begin{tabular}{|c|c|c|c|}
\hline
Method&Train&Test& $\#$ params\\

\hline
EqnDis Poly (2,1) & 0.9109  & 0.850 & 14\\
EqnDis Poly (3,3) & 0.9538 & 0.9435&71 \\

\textbf{EqnDis Fourier (2,1)} & \textbf{0.9737} & \textbf{0.9660} &26\\
EqnDis Fourier (3,1) & 0.9536 & 0.9505 &92\\
EqnDis Fourier (2,2) & 0.9717 & 0.9670 &58\\
\hline
 RNN (5,30)           & 0.9256 & 0.9225 & 8,311\\
 LSTM (2,20)         & 0.9684 & 0.9670 & 5,082\\
 CFN (3,20)           & 0.9409 & 0.9275 & 5,001\\
 NCDE (32,32-1)    & \textbf{0.9786} & \textbf{0.9720} & 1,221\\
 \hline
 \end{tabular}
\caption{ A comparison of the accuracy and number of parameters for various methods on the synthetic dataset based on the forced Lotka-Volterra equation. Each table presents results trained by input data $x(t)$ and $\dot{x}(t)$ respectively.  See Section \ref{data:LV}.}
\label{table:LV}
 \end{table}
Note that the first block of Table \ref{table:LV} shows results with $x(t)$ as input, while the second block shows results with $\dot{x}(t)$ as input.  Comparing these two blocks in Table \ref{table:LV}, we see that across all dictionaries and hyperparameters, the NAED method performs better with $\dot{x}(t)$ as input.  Note that the NAED method with Fourier dictionary yields similar or better results than other methods regardless of whether $x(t)$ or $\dot{x}(t)$ is used as input.

%-------------------------------------------------------------------
\subsubsection{Stochastic gated partial diffusion equation}
\label{s:DiffEq}
Consider the one-dimensional stochastic gated partial diffusion equation \cite{SwitchDiff},
\begin{align*}
& u_t(z,t) = \kappa u_{zz}(z,t) 
&& z \in [0,1], \ t \in [0,1] \\
& u_z(0,t) = 0 && z = 0 \\
& x(t) u(1,t)  + (1-x(t))u_z(1,t) = 0 && z = 1 \\
& u(z, 0) = u_0(z) && t = 0. 
\end{align*}
At $z = 0$, we impose the reflecting (Neumann) boundary condition. 
At $z = 1$, we impose the switching (time-dependent Robin) boundary condition, where for all $t \in [0,1]$, we have $x(t) \in \{0,1\}$, a \emph{switching function}. For the initial condition, we use an approximation to the Dirac delta  $\delta(z-0.5)$, given by 
$u_0(z) = \frac{1}{\sqrt{2\pi \sigma^2}} \exp\left(-\frac{(z-0.5)^2 }{2\sigma^2}\right)$, 
where $\sigma = 0.1$. 
The solution has an interpretation in terms of a particle experiencing Brownian motion on the interval. The probability of finding the particle at time $t$ and position $z$ is given by $u(z,t)$. The initial condition is interpreted as the particles all starting near $z = 0.5$. 
The boundary condition at $z = 1$ has the interpretation that when $x(t) = 1$ a particle leaves the interval when it reaches the boundary 
and when  $x(t) = 0$ the particles are reflected.  
The proportion of particles remaining in the interval at time $t$, referred to as the \emph{survival probability} is given by $ S(t) = \int_0^1 u(z,t) \ dz$. For a given switching function $x(t)$, we assign a  label $y$  based on the survival probability at time $t=1$;
\begin{equation} \label{e:HeatClass}
y = \begin{cases}
(1,0) & S(1) < \frac{1}{2} \\
(0,1) & S(1) \geq \frac{1}{2}
\end{cases}.
\end{equation}
The classification problem seeks the mapping from the switching function $x(t)$ to the binary class $y$. 

We generate a synthetic dataset for this problem with $8000$ training examples and $2000$ testing examples as follows. To generate each switching function $x(t)$ we  choose an integer, $q$, between zero and ten uniformly. We then randomly select $q$ times in the interval $[0,1]$ and starting with $x(t) = 0$, we set $x(t)$ to alternate between 0 and 1 at these times. For each switching function, $x(t)$, we approximately solve the heat equation for $u(z,t)$ as follows. We apply a forward difference in time and a second-order central difference scheme for the space derivative. We use a spatial discretization size of $dx = 0.05$ and temporal step size of $dt = 0.01$.  
To obtain roughly balanced class sizes, we choose the diffusion coefficient to be $\kappa = 0.165$. 
For this choice of parameters, the CFL condition $ \frac{\kappa \ (dt) }{ (dx)^2} \approx 0.66 < 1$ is satisfied, so the numerical method is stable.  The solution at time $t=1$ is used to define the label $y$ as in \eqref{e:HeatClass}.

\begin{table}[t!]
\center
\begin{tabular}{|c|c|c|c|}
\hline
Method&Train&Test& $\#$ params\\
\hline
EqnDis Poly (2,1) & 0.9203 &  0.9155 & 14\\
\textbf{EqnDis Fourier (2,1)} & \textbf{0.9582 }& \textbf{0.9570 }& 26\\
EqnDis Fourier (2,2) & 0.9523 & 0.9515 &58\\
\hline

RNN (3,10)  & 0.9550  & 0.9570 & 146 \\
LSTM (1,5)  & \textbf{0.9805}  & \textbf{0.9799} & 146\\
CFN (2,3)   & 0.9440  & 0.9309 & 88 \\
NCDE (32,32-1)    & 0.9785 & 0.9750 & 1,221\\
\hline

 \end{tabular} 
 \label{table:Heat}
 \caption{A comparison of the accuracy and number of parameters for various methods on the synthetic dataset based on the  stochastic gated diffusion equation. See Section \ref{s:DiffEq}.}
 \end{table}

A comparison of the accuracy of various methods is given in Table \ref{table:Heat}. As shown in the first block of Table \ref{table:Heat}, the proposed NAED method works well on this  dataset generated using a partial diffusion equation. Among the several choices of entries for the dictionary, we achieve the best accuracy with the Fourier (2-1) dictionary. The NAED method provides comparable accuracy to other methods with substantially fewer parameters.

%---------------------
\subsection{Synthetic dataset with noise}\label{data:noise}
In this section, we train the sparse NAED method (see Section \ref{s:sparsity}) and show the robustness of this method on a noisy dataset. To generate the noisy data, we contaminate the forced harmonic oscillator input/forcing $x(t)$ from Section \ref{data:OSC} with noise:
\[
\tilde{x}(t) = x(t) + \eta(t),\qquad \eta(t) \stackrel{iid}{\sim} \mathcal{N}(0,10^{-4})  
\]
where $\eta(t)$ is a Gaussian process, mutually independent for different $t$.  Note that noise is added on the original data $ x(t) \in [-5.8,5.6]$ for $t \in [0,T]$.

 \begin{table}[t!]
 
 \center
\begin{tabular}{|c|c|c|c|}
\hline
Methods &Train&Test& \# nnz params\\
\hline
Poly (2,1) & 0.7580 & 0.7505 & 6\\
Sparse Poly (2,1) & 0.7618 & 0.7605 &3 \\
\hline
Fourier (2,1) & 0.9192 & 0.9155 & 18 \\
Sparse Fourier (2,1) & 0.9311 & 0.928 & 6\\
\hline
Fourier (2,2) & 0.9523 &  0.9515 &50\\
\textbf{Sparse Fourier (2,2)} & \textbf{0.9670} & \textbf{0.9645} & 16\\
\hline

RNN (2,10)  & 0.9557  & 0.9530 & 341 \\
LSTM (2,10)  & 0.9615  & 0.9595 & 1,342\\
CFN (2,10)   & 0.9230  &0.9180  & 781\\
NCDE (16,16-1)  & \textbf{0.9789}  & \textbf{0.9674} & 595\\
\hline
 \end{tabular}
\caption{A comparison of the accuracy and number of nonzero (nnz) parameters for various methods on the synthetic dataset based on the forced harmonic oscillator with noise. See Section \ref{data:noise}.}
\label{table:noise}
 \end{table}

For the noisy data, we apply the sparse NAED method within a cross-validation loop to select $\lambda$.
For each value of $\lambda \in \{0.01, 0.03,0.05, 0.1,0.5, 1\}$, and within each fold of $5$-fold cross-validation, we train with Algorithm \ref{alg:sparsity} until convergence.  We then choose $\lambda$ to minimize the cross-validation test error.  The last column of the Table \ref{table:noise} records the number of non-zero entries in the trained $\beta$. As shown in each block of Table \ref{table:noise}, the performance of the sparse NAED method tends to be slightly better than competing methods.  In particular, the sparse Fourier (2-2) method achieves the best test error with substantially fewer parameters than competing RNN methods. Note that in this synthetic example, the underlying dynamical system does possess a sparse representation.  

%--------------------
\subsection{UCR archive datasets}\label{data:UCR}
 
\begin{figure}[t!]
\centering
\includegraphics[width=.9\linewidth]{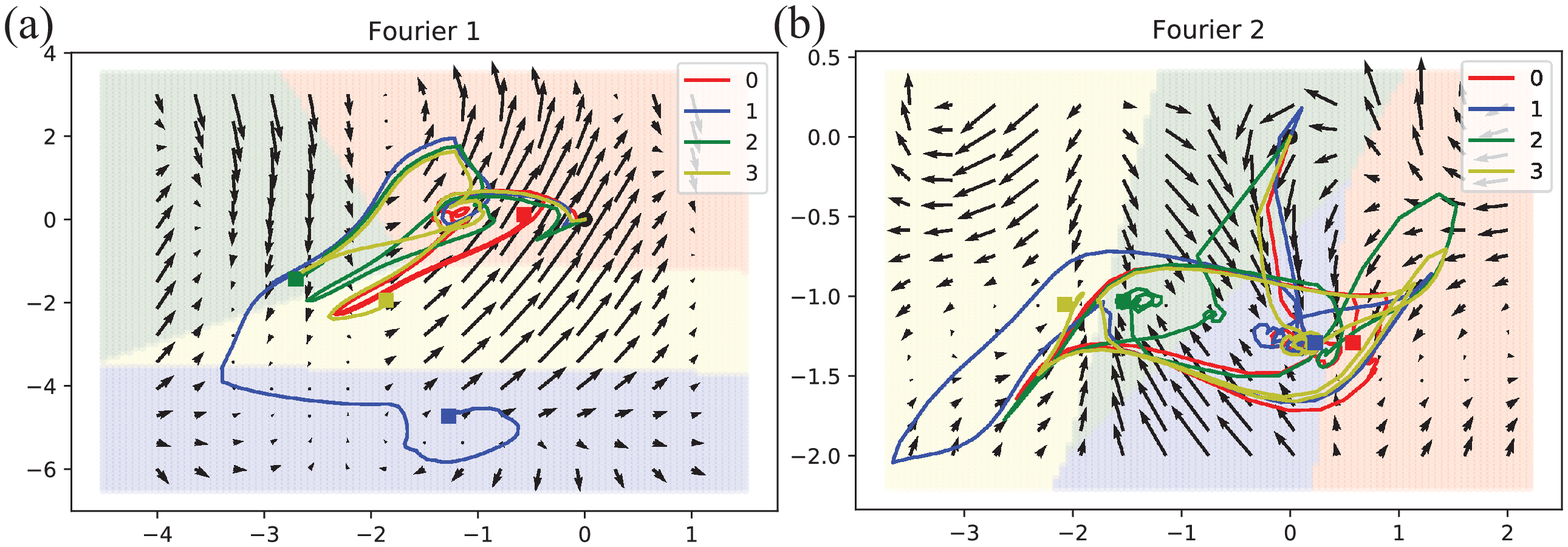}
\includegraphics[width=.9\linewidth]{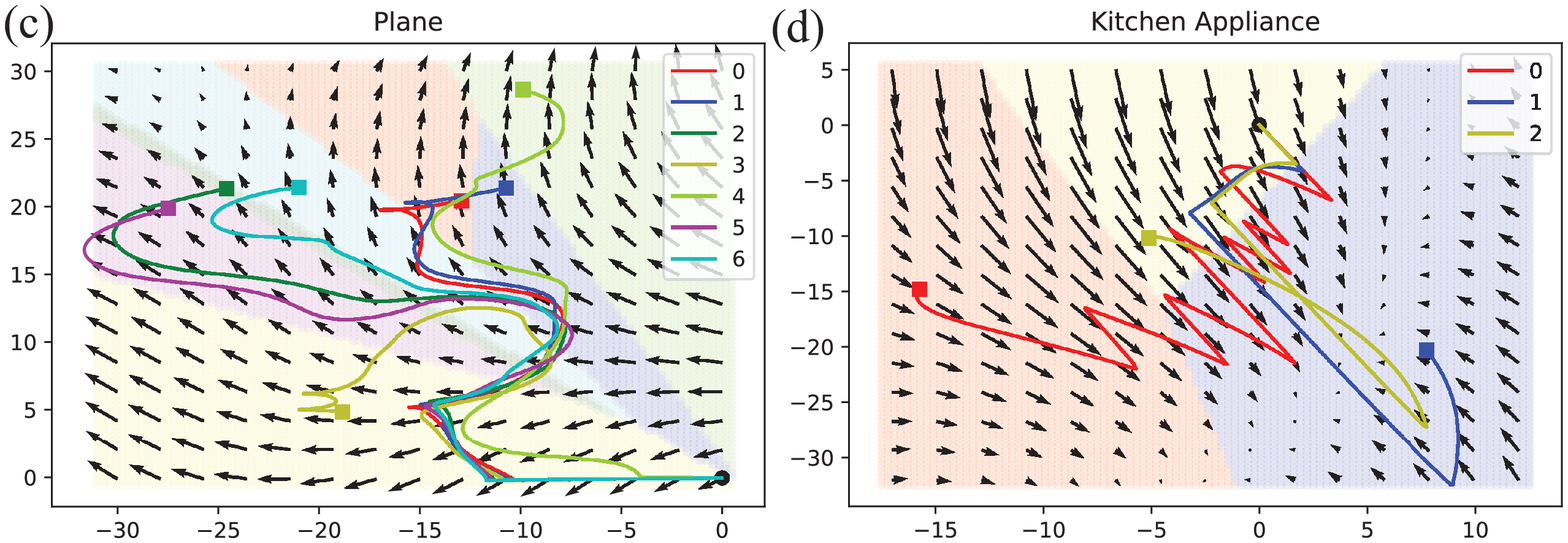}
\caption{In four subplots, labeled {\bf (a)}--{\bf (d)}, we plot the vector field $h \mapsto \beta \Xi(h)$  in \eqref{e:model} trained on different UCR archive datasets.  In each plot, example solution trajectories for each class are displayed and the classification partition is colored. 
{\bf (a)} \textit{Two Patterns} dataset using Fourier dictionary with $K=1$. 
{\bf (b)} \textit{Two Patterns} dataset using Fourier dictionary with $K=2$. {\bf (c)} \textit{Plane} dataset using Fourier dictionary with $K=1$. 
{\bf (d)} \textit{Kitchen Appliance} dataset using sparse-Fourier dictionary with $K=1$ and $\lambda=0.03$. }
\label{fig:UCI}
\end{figure}

In this section, we compare the proposed NAED method with other algorithms on a few univariate time series datasets from the UCR archive \cite{UCR}. For the NAED method, we present the most accurate model by varying the candidate functions of dictionary and cut-off values for sparsity. For the Neural CDE method, we used either 32 or 64 hidden channels; in this method, the vector field is represented using a feedforward neural network with one hidden layer with either 64 or 128 units. The total number of parameters is reported in  \ref{table:UCR}. The results are summarized in Table \ref{table:UCR}. In Figure \ref{fig:UCI}, we show an example trajectory for each class and use colored partitions to denote the classification regions and decision boundaries.

\begin{table}[t!]
\begin{center}
\scalebox{0.8}[1.0]
{\begin{tabular}{|c|c|c||c|c|c|c||c|}
\hline
& Dataset&& RNN & LSTM & CFN & NCDE &~ NAED ~\\ 
\hline
\parbox[t]{2mm}{\multirow{7}{*}{\rotatebox[origin=l]{90}{\qquad UCR archive \qquad\qquad}}} 
&\textbf{Two Patterns}\multirow{6}{*}{ }&test& 0.7630& \textbf{1.0000} &0.9900  &0.8420 &0.9760\\\cline{3-8} 
 & train/test : $1000$/$4000$  &train& 0.7473&\textbf{1.0000} &1.0000 &0.8330 &0.9815 \\\cline{3-8}
  & $4$ classes &   info      &  (5-24-5,428) & (1-35-5,324) & (3-20-5,064)& (64-128-17,030) & (2-1-32)\\\cline{2-8} 
  
 &\textbf{Plane} \multirow{6}{*}{ }&test& 0.7048 &0.4762  & 0.4000 &\textbf{0.8571} &0.7714\\\cline{3-8} 
 &  train/test : $105$/$105$ &train& 0.7429 &0.4952 & 0.5524&\textbf{0.8095} &0.7523 \\\cline{3-8}
 &  $7$ classes& info &  (5-10-3,867) & (5-10-3,917) & (5-20-9,205) & (32-64-8,905)  & (2-1-41) \\\cline{2-8}   
 
  &\textbf{Kitchen Appliance} \multirow{6}{*}{ }&test& 0.5973 & 0.6027 &0.5760 &  0.5306   &\textbf{0.6133}\\\cline{3-8} 
 &train/test : $375$/$375$&train&0.6027 & 0.5813&0.5467& 0.5040 &\textbf{0.6053}
 \\\cline{3-8}
  & $3$ classes & info &  (5-10-993) & (5-10-3,873) & (2-5-228) & (64-64-8,645)& (2-1-29)  \\\cline{2-8}   
  
    &\textbf{Computer}\multirow{6}{*}{ }&test&  0.5800& \textbf{0.6640} & 0.6199& 0.6520 & 0.6599\\\cline{3-8} 
 & train/test : $250$/$250$ &train&  0.5960 & 0.6280&0.6199& \textbf{0.6800} &0.6200\\\cline{3-8} 
  &$2$ classes &   info      &  (1-5-47) & (1-3-68) & (2-2-45) &  (64-128-16,835) & (2-2-58) \\\cline{2-8}   
 
    &\textbf{FordB}\multirow{6}{*}{ }&test&  0.6099&0.4987  &0.5173 &\textbf{0.6185}  &0.5259\\\cline{3-8} 
 & train/test : $3636$/$810$  &train& \textbf{0.7032}&0.5105  & 0.5732 & 0.6468 & 0.5500\\\cline{3-8}
  & $2$ classes&   info      &  (2-2-21) & (1-2-35) & (2-3-88) &(64-128-16,835) &(2-1-25) \\\hline
\end{tabular}}
\end{center}
\caption{ A comparison of the accuracy and number of parameters for various methods on five UCR archive datasets. See Section \ref{data:UCR}. }
\label{table:UCR}
\end{table}

The \textit{Two Pattern} dataset is synthetically generated and has 1000 training and 4000 test samples. There are four balanced classes and the sequence length for all samples is 128. As shown in Table \ref{table:UCR}, the best accuracy is obtained with a Fourier dictionary. Compared with other methods, the NAED method provides slightly lower accuracy but is still close to $100\%$ on both train and test data.  

The \textit{Plane} dataset contains outlines of airplanes measured by a sensor. The classification problem is to distinguish the type of airplane where there are seven airplane shape classes: Mirage, Eurofighter, F-14 wings closed, F-14 wings opened, Harrier, F-22 and F-15.  There are 105 instances in both the training and test sets, each having length $144$.
As presented in Table \ref{table:UCR}, the NAED method (with Fourier 1 dictionary) surpasses the test accuracy of the RNN, LSTM, and CFN.  It does this even with $100$-$200$ times fewer parameters than these competing methods.  The NCDE method is the best on this dataset; it has over $200$ times more parameters than NAED. This dataset shows that NAED works well on a multiclass classification problem.

The \textit{Kitchen Appliance} dataset is behavioral data recorded from 251 households and measured by a device in two-minute intervals over a month. Note that each series has length $720$. This problems classifies how consumers use electricity within their home, so there are three classes: Kettle, Microwave and Toaster. This data contains $375$ instances in the training and test sets. In Table \ref{table:UCR}, NAED with a sparse-Fourier1 dictionary returns the best accuracy on this dataset with only $29$ parameters. Here, the cutoff value is set to $\lambda = 0.03$ and two entries of $\beta$ are dropped to zero. 

The \textit{Computer} dataset consists of 250 train and test instances for a consumer's electricity usage behavior in a home. Each sample consists of recordings made every two minutes over a month so that total length is 720. 
There are two classes: Desktop and Laptop. 
According to Table \ref{table:UCR}, the best accuracy is obtained by the LSTM method.  NAED with sparse-Fourier2 dictionary and cutoff value $\lambda = 0.05$ nearly matches the LSTM's accuracy.  The imposed sparsity condition replaces $16$ entries in $\beta$ with zero; consequently, the trained vector field is relatively simple and interpretable.

The \textit{FordB} dataset contains 3636 training and 810 test instances.  Each instance consists of 500 measurements of engine noise together with a label. The classification problem is to diagnose the existence of certain symptoms in the automotive subsystem, so there are two classes. Note that the training data were collected in typical conditions while test data were collected under noisy conditions. Hence the FordB dataset forces the classifier to generalize from clean to noisy data.  Here, the NAED method is learned using the sparse-Fourier1 dictionary and thresholded by $\lambda = 0.05$. 

For the \textit{Kitchen Appliance} dataset, NAED achieves the best test set results; for the remaining four datasets, NAED's parameter count is on average $>200$ times less than that of the method with the best test set performance.  
For the first three datasets considered in Table \ref{table:UCR}, NAED is the only method that achieves competitive test set results with a small number of parameters.  For \textit{Computer}, the parameter counts for NAED and LSTM are similar.  For \textit{FordB}, RNN performs surprisingly well with a low parameter count.  Based on the RNN results here, we conjecture that NAED underfits this dataset; a more scalable implementation of the NAED method would enable us to explore larger values of the dimension of $h$ and the largest Fourier multiplier $K$.

As we described in Section \ref{sec2}, the NAED method learns a representation of the underlying vector field based on a prespecified dictionary. With polynomial or harmonic basis functions, these vector fields can be approximated using only a few terms.  By promoting sparsity, Algorithm \ref{alg:sparsity} can further enhance parsimony.  As shown in experiments, competing methods require at least $2$ times and up to $500$ times the number of parameters required by NAED.

 %%%%%%%%%%%%%%%%%%%%%%%%%%%%%%%%%%%%%
\section{Discussion} \label{sec4}
In this paper, we developed a framework for analyzing time signals based on non-autonomous dynamical systems. 
A time signal, $x(t)$, is interpreted as a forcing function for a dynamical system \eqref{e:model} that governs a time-evolving hidden variable, $h(t)$. 
As in equation discovery, the dynamical system is represented using a dictionary of prespecified candidate functions and the coefficients are learned from data.
We refer to the resulting model as non-autonomous equation discovery (NAED). This framework is applied to the time signal classification problem, where the hidden variable, at a final time, $h(t=T)$, is used to make a prediction via the composition of the softmax function and an affine function. Using a cross-entropy loss function, we train the NAED model using a gradient based optimization method, where the gradients are efficiently computed using the adjoint method; see Theorem~\ref{p:grad}.

Through a variety of experiments---on both synthetic and real datasets---we demonstrated that the NAED method achieves accuracy that is comparable to RNN, LSTM, CFN and NCDE methods on binary and multi-class classification problems; see Section \ref{sec3}.  Note that \cite{Lyons2020} shows that NCDE itself outperforms other RNN architectures, including continuous-time/ODE-like GRU models \cite{Moreau2019,Park2019} and a method that merges an RNN with a neural ODE \cite{Duvenaud2019}.
The NAED method generally requires far fewer parameters than neural network-based methods and the number of parameters can further be reduced by using a sparse version of the algorithm; see Algorithm \ref{alg:sparsity}. We also show in Section \ref{data:UCR} that sparsity improves  the trainability of the method and its robustness to noise in the data. Finally, by construction, our method is interpretable using the theory of dynamical systems. For example, using phase plots, we can visualize the trajectories of the underlying dynamical system and how they navigate the decision boundaries between classes. 

Since our model is built on dynamical systems, we can generate synthetic data from a dynamical system and then pose the \emph{inverse problem} of trying to recover the ground-truth system from the data. 
For a synthetic dataset based on the forced harmonic oscillator (Section \ref{data:OSC}), we showed that the NAED method for classification is not generally identifiable, \ie, the method does not always recover the ground-truth parameters. However, in the case of a linear dictionary, we recover the ground-truth parameters up to conjugation by an orthogonal matrix. 

There are a variety of natural future directions for this work. Since the NAED method is built on dynamical systems, we could use dynamical systems theory to further analyze a particular trained NAED model. For example, one could use stability theory to further sharpen and generalize the misclassification estimates in Theorem~\ref{p:stability} and \ref{p:stabilityWiener}. To enhance the method's ability to deal with noisy time signals, one could combine the NAED method with filtering methods (\eg, the Kalman filter).
Since we interpret time signals as continuous objects and discretize within the method (the optimize-then-discretize approach), multi-scale methods could be used in training. 
A slight generalization of the model would be to let  $B$ in \eqref{RHS} be a parameterized operator,
$B  = \sum_{k=0}^K B_k \partial_t^k$,
where $B_k \in \mathbb R^{m \times n}$ are unknown coefficients.  In the forced Lotka-Volterra equations (Section~\ref{data:LV}), we considered using as forcing either $x$ or $\dot x$ and this generalization would avoid this.
Finally, the NAED framework developed here could be applied to other time signal analysis tasks, such as  prediction and forecasting, classification, segmentation, and denoising.

\begin{acknowledgements}
We would like to thank Dong Wang and Rebecca Hardenbrook for helpful discussions in the early stages of this work. 
\end{acknowledgements}
 
\section*{Conflict of interest}
The authors declare that they have no conflict of interest.

%\clearpage
%\bibliographystyle{spmpsci}
%\bibliography{references}

\clearpage
\appendix 
\normalsize
\section{Proofs} \label{s:App}
\begin{proof}[Theorem~\ref{prop:exun}.]
Let $\Phi(h,x(t))$ be rewritten as $\Gamma(h,t) = \Phi(h,x(t))$. For some $r>0$ and $a>0$, define  $ B_r = \{\|h-h_0
\|\leq r\}\subset K, I_a = \{|t|\leq a\}$. Since $x$ is continuous in time, there exists a constant $M>0$ such that 
$$M = \max_{(h,t) \in  B_r\times I_a} \ \|\Gamma(h,t)\|.$$
 Also for every $t \in I_a$, $h \mapsto \Gamma(h,t)$ satisfies the local Lipschitz condition on $K$:  for every $h_1, h_2 \in B_r$,
\begin{multline*}
\|\Gamma(h_1,t) - \Gamma(h_2,t)\| 
\ = \ \|\beta(\Xi(h_1)- \Xi(h_2))\|
\ \\ \leq \  \|\beta \| \ \|(\Xi(h_1)- \Xi(h_2))\|
\ \leq \  d\mathcal{L} \|\beta\| \ \|h_1-h_2\|.
\end{multline*}
From the existence/uniqueness theorem in ordinary differential equations  (see, \eg, \cite[Theorem 3.2]{ODE}), there exists a unique solution to \eqref{e:model} on the interval $[-\varepsilon,\varepsilon]$, where $\varepsilon$ is chosen as  $\varepsilon =\min\{a,\frac{r}{M},\frac{1}{2d\|\beta\|\mathcal{L}}\}$. \qed 
\end{proof}

\begin{proof}[Theorem~\ref{p:grad}.]
We introduce the Lagrange multipliers, $\lambda_i \colon [0,T_i] \to \mathbb R^m$, for $i \in [N]$, and the  Lagrangian, 
\begin{align*}
L(\Theta, h_i, \lambda_i) 
&= J(\Theta) + \sum_{i \in [N]} \int_0^{T_i} \lambda^t_i(t) \left( \dot h_i(t) - \Phi(h_i(t),x_i(t)) \right) \ dt \\
\nonumber
&= J(\Theta) + \sum_{i \in [N]} \lambda^t_i(T_i) h_i(T_i) -\int_0^{T_i}  \dot \lambda^t_i(t) h_i(t) + \lambda^t_i(t)\Phi(h_i(t),x_i(t)) \ dt.
\end{align*}
Here, we have used integration by parts to rewrite the Lagrangian. Taking the variation of the Lagrangian with respect to $h_k(t)$ gives
$$
\delta L =  \partial_{h_k(T)} J \delta h_k(T)  
+ \lambda^t_k(T_k) \delta h_k(T_k)
-  \int_0^{T_k} \left( \dot \lambda^t_k(t)  +  \lambda_k^t D_h \Phi  \right)  \delta h_k(t) \ dt ,
$$
where $D_h \Phi = \beta D_h\Xi(h) $ is the Jacobian of $\Phi$ with respect to the $h$. Setting the variation to zero, we find that $\lambda_k(t)$ satisfies the adjoint equation given in \eqref{e:adj}.

The gradients of the objective in  \eqref{e:grad} are then obtained by taking the partial derivatives of the Lagrangian with respect to the unknown parameters, $\Theta = \{\beta,B,A,b\}$. The gradient with respect to $\beta$ and $B$ are given by
\begin{align*}
\nabla_\beta J= \nabla_\beta L &= -\sum_{i \in [N]} \int_0^{T_i} \lambda_i(t)\Xi(h_i(t))^t   \ dt \\
\nabla_B J= \nabla_B L &= - \sum_{i \in [N]} \int_0^{T_i} \lambda_i(t) x_i(t)^t   \ dt.
\end{align*}
For the cross-entropy loss function in \eqref{e:Loss}, a short computation shows that 
\begin{align*}
\nabla_A J &=  - \frac{1}{N}  \sum_{i\in [N]}\left( y_i -\sigma(Ah_i(T_i)+b)\right) h_i(T_i)^t \\
 \nabla_b J  &=  - \frac{1}{N}  \sum_{i\in [N]} (y_i -\sigma(Ah_i(T_i)+b)).
\end{align*}
Combining these results concludes the proof. \qed 
\end{proof}

\begin{proof}[Theorem~\ref{p:stability}.]
In the NAED method with dictionary $\Xi$, the unperturbed and perturbed hidden variables, $h$ and $\tilde h$ satisfy
\begin{align*}
\frac{d}{dt}  h &= \beta \Xi(h) + B  x \\
\frac{d}{dt}  \tilde h &= \beta \Xi( \tilde h) + B \tilde x, 
\end{align*}
with $h(0) = \tilde h (0) = h_0$. 
Let $\mathcal L$ denote the Lipschitz constant for the dictionary $\Xi$. 
Subtracting these equations, we  estimate
\begin{equation*}
|h(t) - \tilde h(t)|  \leq  \int_0^t \mathcal L \|\beta\|  \ |h(\tau)-\tilde h(\tau)| ~ d\tau  
+ \int_0^t \|B\| \ 
|\eta(\tau)|~d\tau.
\end{equation*}
Since 
$\int_0^t \|B\| \ |\eta(\tau)|~d\tau$ is a non decreasing function, Gronwall's inequality yields 
\begin{equation*}
|h(T) - \tilde h(T)|  \leq  \|B\| \left(\int_0^T |\eta(\tau)|~d\tau\right) ~e^{\mathcal L T \|\beta\|} = C\int_0^T |\eta(\tau)|~d\tau,
\end{equation*}
where $C = \|B\| e^{\mathcal L T \|\beta\|}$.
The softmax prediction function in \eqref{e:softmax} is Lipschitz continuous with constant that we denote by $L_{\sigma}$. We have 
\begin{equation}\label{e:pred_bound}
     | \mathscr{C}(\tilde x) - \mathscr{C}(x) | \leq L_\sigma | \tilde h (T) - h(T) |
 \leq L  \| \eta \|_{L^1\left([0,T]; \mathbb R^n \right)} , 
\end{equation}
 where $L = L_{\sigma} C$, as desired. \qed 
\end{proof}

\begin{proof}[Theorem~\ref{p:stabilityWiener}.]
In the NAED method with dictionary $\Xi$, the unperturbed and perturbed hidden variables, $h$ and $\tilde h$ satisfy
\begin{align*}
h(t) &= h_0 + \int_0^t \beta \Xi(h(\tau)) \, d\tau + B \int_0^t x(\tau) \, d\tau \\
\tilde{h}(t) &= h_0 + \int_0^t \beta \Xi( \tilde h(\tau) ) \, d\tau + B \int_0^t x(\tau) \, d\tau + B \int_0^t dW_{\tau}
\end{align*}
with $h(0) = \tilde h (0) = h_0$. 
Subtracting these equations, we first obtain
\[
|\tilde h(t) - h(t)| = \left| \int_0^t \beta \left[ \Xi(h(\tau))  - \Xi(\tilde h(\tau)) \right]d \tau + B W_t \right|.
\]
Let $\mathcal L$ denote the Lipschitz constant for the dictionary $\Xi$.  We estimate
\begin{align*}
|\tilde h(t) - h(t)|   &\leq  \int_0^t \mathcal L \|\beta\|  \ |h(\tau)-\tilde h(\tau)| ~ d\tau  
+  \|B\| |W_t| \\
&\leq  \int_0^t \mathcal L \|\beta\|  \ |h(\tau)-\tilde h(\tau)| ~ d\tau  
+  \|B\| \sup_{0 \leq s \leq t} |W_s|
\end{align*}
Continuity of $W_t$ implies continuity of $\sup_{0 \leq s \leq t} |W_s|$.  Note that $\sup_{0 \leq s \leq t} |W_s|$ is non-decreasing.  Hence Gronwall's inequality yields 
\begin{equation*}
|\tilde h(T) - h(T)|  \leq  \|B\| \sup_{0 \leq s \leq T} |W_s| e^{\mathcal L T \|\beta\|} = C \sup_{0 \leq s \leq T} |W_s|,
\end{equation*}
where $C = \|B\| e^{\mathcal L T \|\beta\|}$.   We combine this with the Lipschitz bound on softmax:
\begin{equation}\label{e:pred_bound2}
     | \mathscr{C}(\tilde x)  - \mathscr{C}(x) | \leq L_{\sigma} | \tilde h (T) - h(T) |
 \leq L  \sup_{0 \leq s \leq T} |W_s|,
\end{equation}
 where $L = L_{\sigma} C$ as before.  The remaining estimates can be derived from the density computed in \cite[\S2.8A]{Karatzas}; for clarity, we provide a self-contained treatment.  Let $B_t$ denote the Wiener process in $\mathbb{R}$, and let $\tau_z = \min\{ t : B_t = z \}$, a first passage time.  Then note that
 \[
 P(B_t \geq z) =   \underbrace{P(B_t \geq z \, | \, \tau_z \leq t)}_{\text{I}} P(\tau_z \leq t) + \underbrace{P(B_t \geq z \, | \, \tau_z > t)}_{\text{II}} P(\tau_z > t)
 \]
By symmetry of $B_t$, term I is $1/2$; by continuity of $B_t$, term II is $0$.  Hence $P(\tau_z \leq t) = 2 P(B_t \geq z) = \operatorname{erfc}(z (2t)^{-1/2})$ where  $\operatorname{erfc}$ is the complementary error function.  Now using the reflection principle, we have
\begin{align*}
P \biggl( \sup_{0 \leq s \leq T} |B_s| \geq z \biggr) &\leq 2 P \biggl( \sup_{0 \leq s \leq T} B_s \geq z \biggr) \\
 & \leq 2 P(\tau_z \leq T) \\ &\leq  2 \operatorname{erfc}(z (2T)^{-1/2}) \\
 &\leq 2 e^{-z^2/(2T)}.
\end{align*}
Let $W_{t,j}$ denote the $j$-th coordinate of $W_t$; each $W_{t,j}$ is an independent one-dimensional Wiener process.  With $|w|_p$ denoting the $p$-norm of the vector $w \in \mathbb{R}^d$, we have $|w| = |w|_2 \leq d^{1/2} |w|_{\infty}$.  Putting these facts together, we estimate
\begin{align*}
P \biggl( \sup_{0 \leq s \leq T} |W_s| \geq z \biggr) 
    &\leq P \biggl( \sup_{0 \leq s \leq T} |W_s|_{\infty} \geq z d^{-1/2} \biggr) \\
    &\leq P \biggl( \sup_{1\leq j \leq d} \sup_{0 \leq s \leq T} |W_{s,j}| \geq z d^{-1/2} \biggr) \\
    &\leq d P \biggl( \sup_{0 \leq s \leq T} |B_s| \geq z d^{-1/2} \biggr) \\
    &\leq 2 d e^{-z^2/(2 d T)}.
\end{align*}
\end{proof}
Combining this with (\ref{e:pred_bound}) yields the conclusion of the theorem. \qed
\end{document}